\documentclass{ifacconf}

\usepackage{amsfonts}
\usepackage{amsmath}
\usepackage{graphicx}
\usepackage{natbib}
\usepackage{algorithm}
\usepackage{algpseudocode}
\usepackage{subcaption}
\usepackage{setspace}

\begin{document}
\begin{frontmatter}

\title{Neural Distance-Guided Path Integral Control for Tractor-Trailer Navigation} 

% \thanks[footnoteinfo]{This work is not sponsored by any grant.}

\author[1]{Peng Wei} 
\author[2]{Chen Peng} 
\author[1]{Stavros Vougioukas}

\address[1]{Department of Biological and Agricultural Engineering, University of California Davis,
   Davis, CA 95616 USA (e-mail: \{penwei, svougioukas\}@ucdavis.edu).}
\address[2]{ZJU-Hangzhou Global Scientific and Technological Innovation Center, Zhejiang University, Hangzhou, 311215, China (e-mail: chen.peng@zju.edu.cn)}

\begin{abstract}                % Abstract of 50--100 words
% Tractor-trailer systems
Autonomous and safe navigation of tractor–trailer systems requires accurate, real-time collision avoidance and dynamically feasible control, particularly in cluttered and complex agricultural environments. This is challenging due to their articulated, deformable geometries and nonlinear dynamics. Traditional methods oversimplify vehicle geometry or rely on precomputed distance fields that assume a known map, limiting their applicability in dynamic, partially unknown environments. To address these limitations, we propose a geometric neural encoder that provides fast and accurate distance estimates between the full tractor–trailer body and raw LiDAR perception, enabling real-time, map-free geometric reasoning. These learned distances are integrated into a Model Predictive Path Integral (MPPI) controller, allowing the system to incorporate true articulated geometry directly into its  cost evaluation and enabling more responsive navigation in challenging agricultural settings. Simulation results demonstrate that the proposed framework generates dynamically feasible and safe trajectories for navigating tractor–trailer systems in cluttered and complex environments.
\end{abstract}


\end{frontmatter}
%===============================================================================

\section{Introduction}
\label{sec:introduction}
Articulated tractor–trailer vehicles are widely used in agricultural operations for transporting crops, towing implements, and maneuvering within orchards, vineyards, and open farmlands. Compared with single-body agricultural vehicles, articulated systems introduce additional complexity due to multi-body coupling, off-axle hitching, articulation constraints, and the potential for unstable reversing behavior. These challenges are further amplified in cluttered agricultural settings that feature narrow spaces, static obstacles such as tree rows and storage bins, and dynamic objects including farm workers and other machinery. Ensuring safe and reliable navigation in such environments is therefore essential for enabling fully autonomous operation.

Prior research has examined the stability and controllability of articulated tractor–trailer systems used not only in agricultural settings but also in industrial and heavy-duty transport applications, highlighting issues such as jackknifing and degraded reversing performance \citep{david2014control}. Various path-tracking and optimal control approaches have been explored, including optimal control formulations for headland turning \citep{oksanen2004optimal}, hybrid control strategies with global attractors \citep{altafini2002hybrid}, model predictive control for coordinated tractor–trailer regulation \citep{backman2010nonlinear}, and safety-governed articulation control for reverse motion \citep{hejase2018constrained}. Additional efforts focus on minimizing swept path during reversing \citep{liu2019minimum} or applying reinforcement learning to obtain robust control policies \citep{kang2024deep}. However, these methods primarily address path following and control tasks in simple geometric environments and do not explicitly handle dense obstacle fields from the environment.

Trajectory planning methods that explicitly account for environmental obstacles have also advanced for articulated systems. The progressively constrained nonlinear programming framework in \cite{li2019tractor} and the hierarchical optimization strategy in \cite{li2021optimization} demonstrate safe navigation capabilities, but both approaches rely on accurate prior maps and remain computationally demanding. More recent trajectory deformation method by \cite{xu2025tracailer} offers improved efficiency, yet require maintaining an Euclidean Signed Distance Field (ESDF), which is costly to update in dynamic or occluded agricultural environments. 

% A comprehensive review of planning and control techniques for tractor–trailer systems is provided in \cite{yang2025review}.

Navigating articulated vehicles in cluttered environments requires accurately computing distances between the vehicle body and surrounding obstacles. This is particularly challenging for tractor–trailer systems, whose articulated geometries make collision checking expensive. Moreover, simple occupancy checks are insufficient for safe navigation, as they provide only binary collision information and cannot indicate how close the vehicle is to nearby obstacles—information that is essential for predictive avoidance and smooth control. Simplified representations such as circles or inflated bounding boxes reduce computation but produce inaccurate distance estimates and can make navigation unnecessarily conservative or even infeasible in tightly constrained environments. Safe-corridor–based methods \citep{wei2025efficient} depend on a known map and can become overly restrictive when the calculated corridor is conservative. ESDF–based approaches \citep{xu2025tracailer} provide better distance information without explicit corridor construction, but maintaining an ESDF in dynamic or partially occluded agricultural environments is costly. Optimization-based Collision Avoidance (OBCA) \citep{zhang2020optimization} enables exact polygon-to-polygon distance queries through a duality-based formulation, but solving the associated optimization problem when many obstacles are present is too slow for real-time navigation.

These limitations highlight the need for a map-free, accurate, and computationally efficient geometric distance model that operates directly on raw sensor data. To address this need, we develop a geometric neural encoder that computes fast, accurate minimum distances between a tractor–trailer body and raw LiDAR point clouds. Using the dual optimization formulation of point-to-polygon distance, the encoder learns to predict dual variables efficiently, enabling real-time geometric reasoning without prior maps or geometric simplifications.

For motion planning, we adopt the Model Predictive Path Integral (MPPI) control framework, a sampling-based optimal control method well suited for the nonlinear articulated dynamics of tractor–trailer systems and the discontinuous collision penalties arising from raw distance queries. MPPI has demonstrated strong performance in aggressive, real-time autonomous driving tasks when implemented on modern GPUs \citep{williams2016aggressive}. By incorporating neural distance estimates directly into the MPPI cost evaluation, the controller can react responsively to LiDAR observations and generate safe, dynamically feasible trajectories through narrow and cluttered agricultural fields.

% Contributions
This work makes the following three contributions:
\begin{enumerate}
\item We develop a geometric neural encoder that provides fast and accurate signed-distance estimates between articulated tractor–trailer geometries and raw LiDAR point clouds, enabling real-time geometric reasoning without requiring a pre-built map.

\item We integrate these distance estimates into a Model Predictive Path Integral framework, enabling perception-driven local planning and collision avoidance for articulated agricultural vehicles operating under nonlinear dynamics and complex cost structures.

\item We evaluate the proposed pipeline in simulations and demonstrate its effectiveness in navigating tractor–trailer systems safely and autonomously through cluttered and complex environments.
\end{enumerate}

% The remainder of this paper introduces the tractor–trailer model, the geometric neural encoder, and the MPPI controller, followed by simulation results and a conclusion outlining future extensions.

%%%%%%%%
\section{Tractor--Trailer Kinematic Model}
\label{sec:kinematic_model}
In this work, we model the tractor--trailer system using a simplified kinematic formulation. Because agricultural machinery typically operates at low speeds in field environments, lateral inertial effects are small, and a no-slip kinematic bicycle model is adopted as a first-order approximation. Although slip and model mismatch can still arise from soft soil, uneven terrain, tight maneuvers, and towing loads, the kinematic model remains widely used in agricultural robotics for its tractability and accurate representation of the tractor–trailer geometric coupling. The trailer considered here is a single off-axle semi-trailer, where the hitch point is located at a distance behind the tractor's rear axle.

Let the state of the system be described by the tractor rear-axle position $(x,y)$, tractor heading $\theta$, and the articulation angle $\phi = \theta_1 - \theta$ between the trailer heading $\theta_1$ and tractor heading $\theta$. Under these assumptions, the kinematic model of the articulated vehicle is
\begin{equation}
\label{eq:kinematic_model}
\left\{
\begin{aligned}
\dot{x} &= v \cos\theta, \\
\dot{y} &= v \sin\theta, \\
\dot{\theta} &= \frac{v}{L_0}\tan\psi, \\
\dot{\phi} &= 
v\left(
   -\frac{\sin\phi}{L_1}
   -\frac{\tan\psi}{L_0}
   -\frac{L_h\,\cos\phi\,\tan\psi}{L_0 L_1}
\right),
\end{aligned}
\right.
\end{equation}

where $v$ is the tractor's longitudinal velocity and $\psi$ is the tractor steering angle. $L_0$ is the tractor wheelbase, $L_1$ is the distance from the hitch point to the trailer axle, and $L_h$ is the hitch offset from the tractor rear axle. The state is $[x,y,\theta,\phi]^T$, and the control input is $[v, \psi]^T$.

Given the tractor pose $(x,y,\theta)$ and articulation angle $\phi$, the trailer heading can be recovered as $\theta_1 = \theta + \phi$. The global coordinate of the trailer axle center $(x_1,y_1)$ is obtained by first computing the hitch position,
\begin{equation}
    x_h = x - L_h\cos\theta, \qquad 
    y_h = y - L_h\sin\theta,
\end{equation}

and then projecting along the trailer axis:
\begin{equation}
    x_1 = x_h - L_1\cos\theta_1, \qquad
    y_1 = y_h - L_1\sin\theta_1
\label{eq:trailer_position}
\end{equation}

\begin{figure}[t]
\centering
    \includegraphics[width=0.7\linewidth]{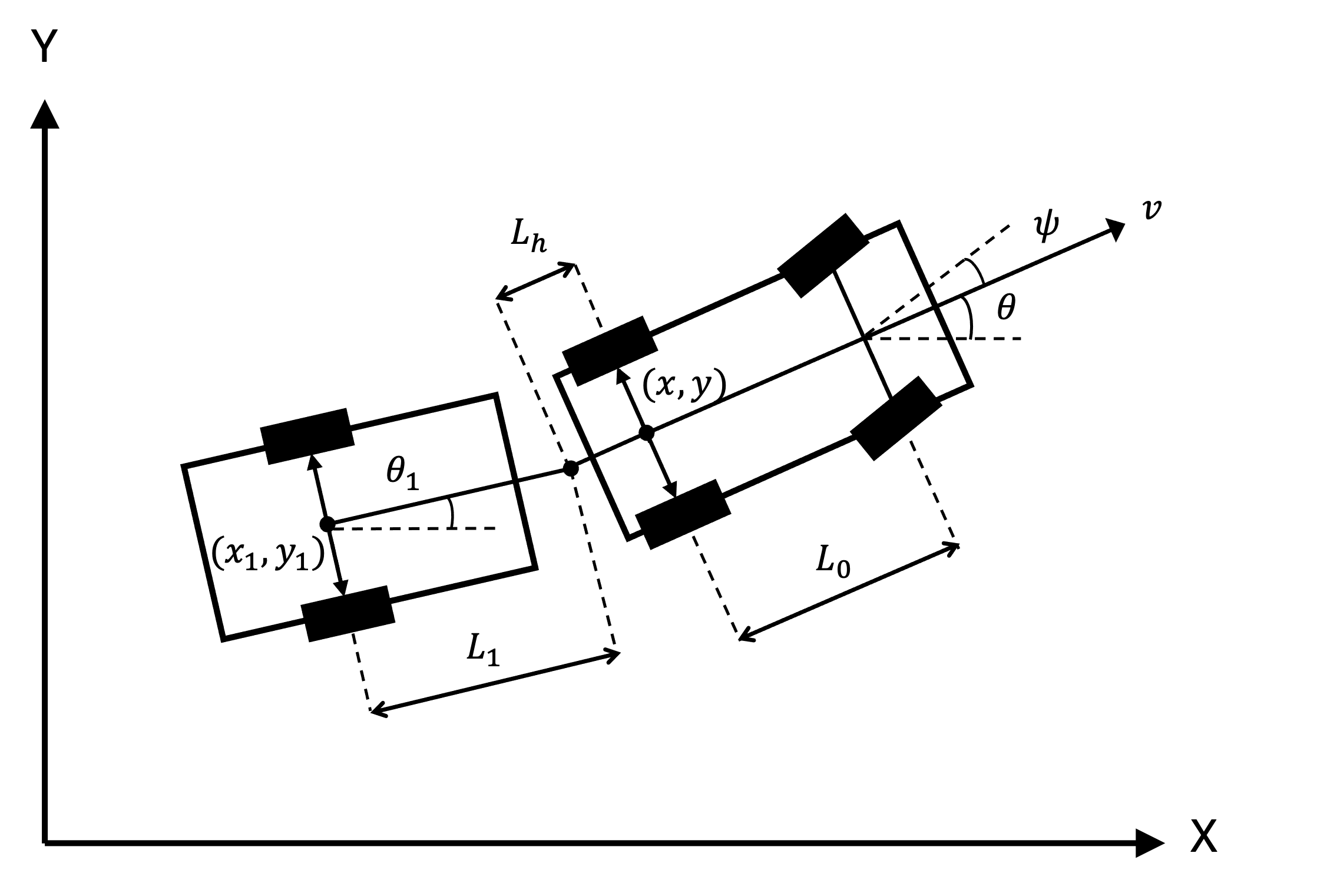}
\caption{Geometric configuration of a tractor–trailer system.}
\label{fig:tractor_trailer_diagram}
\end{figure}

Figure~\ref{fig:tractor_trailer_diagram} illustrates the geometric parameters and coordinate definitions used in this model. Note that the above formulation can be extended to systems with multiple trailers by introducing additional articulation angles and hitch lengths. However, in this work, we focus on a single trail configuration to simplify the analysis and validate the proposed pipeline before extending it to more complex articulated systems.

\section{Geometric Neural Encoder for Efficient Distance Computation}
\label{sec:neural_encoder}
Safe navigation of the articulated tractor–trailer system requires maintaining sufficient clearance between the vehicle body and surrounding obstacles. Traditional methods either approximate the vehicle using simplified geometric primitives or inflate the environmental obstacles to simplify collision checking. Although computationally convenient, these approaches often lead to inaccurate distance estimates and overly conservative navigation, especially for vehicles with complex articulated geometries. To address this limitation, we leverage a geometric neural encoder that accelerates minimum-distance computation directly from raw LiDAR point clouds to the full tractor–trailer body geometry.

Since any nonconvex geometry can be expressed as union of convex sets, we represent the tractor and trailer bodies as collections of convex polygons, $\{\mathbb{V}_{1}, \mathbb{V}_{2}, \dots, \mathbb{V}_{K}\}$, where $\mathbb{V}_{1}, \dots, \mathbb{V}_{k}$ belongs to the tractor body and $\mathbb{V}_{k+1}, \dots, \mathbb{V}_{K}$ belong to the trailer body.  Each convex polygon $\mathbb{V}_i$ in its local body frame is represented as the polytope:
\begin{equation}
    \mathbb{V}_i = \{ \vec{x} \in \mathbb{R}^2 \mid \vec{G}_i \vec{x} \leq \vec{h}_i \}
\end{equation}
where $\vec{G}_i \in \mathbb{R}^{l_i \times 2}$, $\vec{h}_i \in \mathbb{R}^{l_i}$, and $l_i$ is the number of polygon edges. Denote the state of the tractor-trailer system at time $t$ as $\vec{s}_t$, the rotation matrix of the tractor polygon is $\vec{R}(\vec{s}_t) \in \mathbb{R}^{2 \times 2}$ and the corresponding translation is $\vec{t}(\vec{s}_t) \in \mathbb{R}^2$. The occupied region of polygon $i$ in the global frame is then:
\begin{equation}
    \mathbb{E}^{i}_t = \{ \vec{R}(\vec{s}_t) \vec{x} + \vec{t}(\vec{s}_t) \mid \vec{x} \in \mathbb{V}^i \}
\end{equation}
The same transformation applies to all polygons on both the tractor and the trailer, with the rotation matrix $\vec{R}_1$ and translation vector $\vec{t}_1$ for the trailer polygons defined by the pose $(x_1, y_1, \theta_1)$ of the trailer body (formulation omitted here for simplicity). 

Assume the onboard LiDAR sensor reads $M$ obstacle points at time $t$, $\mathbb{O}_t = \{ \vec{o}_t^1, \dots, \vec{o}_t^M \}$, with $\vec{o}_t^j \in \mathbb{R}^2$. The minimum distance from $i$-th polygon $\mathbb{E}_t^i$ to the point cloud at time $t$ is:
\begin{equation}
    \text{dist}(\mathbb{E}_t^i,~\mathbb{O}_t) = \min_{j=1,\dots,M} \{\mathcal{D}(\mathbb{E}_t^i, \vec{o}_t^j)\}
\label{eq:min_distance}
\end{equation}
where $\mathcal{D(\cdot)} \in \mathbb{R}$ is the exact point-to-polygon distance, obtained by solving:
\begin{equation}
\min_{\vec{x}} || \vec{R}(\vec{s_t}) \vec{x} + \vec{t}(\vec{s}_t) - \vec{o}_t^j ||^2, \quad s.t. \quad \vec{G}_i \vec{x} \leq \vec{h}_i
\label{eq:original_ocp}
\end{equation}
By leveraging the strong duality result in \cite{zhang2020optimization}, this distance computation can be equivalently reformulated as the dual problem:
\begin{equation}
\begin{aligned}
    \max_{\vec{\mu}_t^i, \vec{\lambda}_t^i} \quad & {\vec{\mu}_t^i}^T (\vec{G}_i \tilde{\vec{o}}_t^j - \vec{h}_i) \\
    s.t.\quad & {\vec{\mu}_t^i}^T \vec{G}_i + {\vec{\lambda}_t^i}^T \vec{R}(\vec{s}_t) = 0 \\
        & \vec{\mu}_t^i \geq 0,~ || \vec{\lambda}^i|| \leq 1
\end{aligned}
\label{eq:dual_problem}
\end{equation}
where the obstacle point expressed in the polygon frame is $\tilde{\vec{o}}_t^j = \vec{R}(\vec{s}_t)^T (\vec{o}_t^j - \vec{t}(\vec{s}_t))$.  The dual variables have a useful geometric interpretation: $\vec{\mu}_t^i \in \mathbb{R}^{l_i} $ identifies the closest polygon edge(s) and $\vec{\lambda}_t^i \in \mathbb{R}^2 $ encodes the direction of the distance vector. Thus each obstacle point is sparsely associated with its nearest edge(s) on the polygon.

A penalized version of the dual problem in (\ref{eq:dual_problem}) is:
\begin{equation}
\begin{aligned}
    \min_{\vec{\mu}_t^i, \vec{\lambda}_t^i} \quad & {\vec{\mu}_t^i}^T (\vec{h}_i - \vec{G}_i \tilde{\vec{o}}_t^j) + w_p ||{\vec{\mu}_t^i}^T \vec{G}_i + {\vec{\lambda}_t^i}^T \vec{R}(\vec{s}_t) ||^2 \\
    s.t.\quad & \vec{\mu}_t^i \geq 0,~ || \vec{\lambda}^i|| \leq 1
\end{aligned}
\label{eq:dual_problem_penalize}
\end{equation}
where $w_p$ is a sufficiently large penalty coefficient. Due to strong convexity, (\ref{eq:dual_problem_penalize}) can be efficiently solved using inexact block coordinate descent optimization, alternating between updates of $\vec{\mu}_t^i$ and $\vec{\lambda}_t^i$. However, evaluating this optimization for a large number of LiDAR points remains slow. 

To achieve real-time performance while retaining accuracy, we use a geometric neural encoder that learns to approximate the solution of  (\ref{eq:dual_problem_penalize}). As observed in \cite{han2025neupan}, each iteration of the dual optimization consists primarily of linear operations and simple nonlinearities, making it suitable for unrolled neural architectures. 

\begin{figure}[t!]
\centering
    \includegraphics[width=0.7\linewidth]{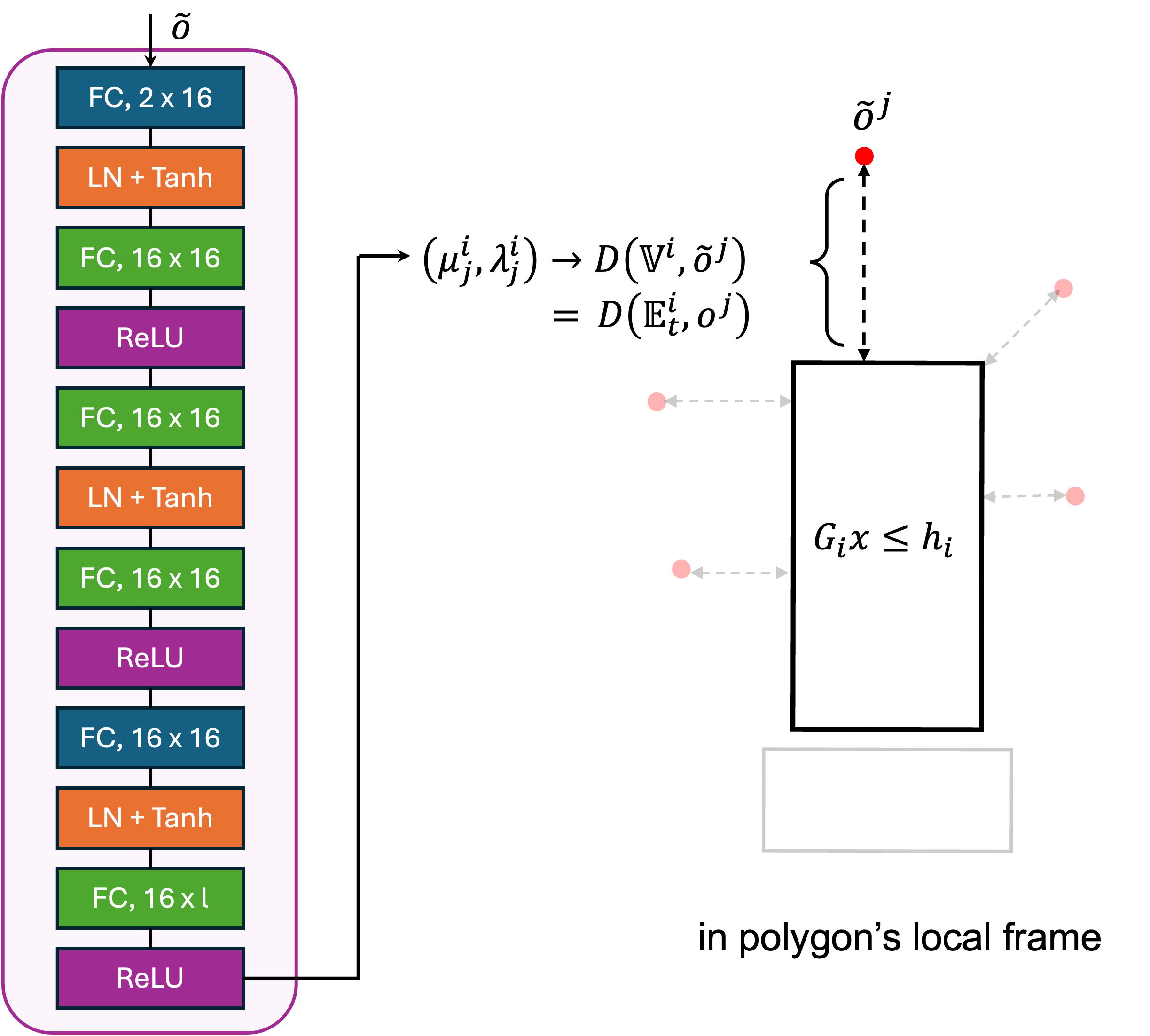}
\caption{The architecture of the geometric neural encoder (left) and the geometric relationship between a convex polygon and transformed obstacle points (right).}
\label{fig:network}
\end{figure}

The architecture of the geometric neural encoder network is shown in Fig.~\ref{fig:network}. The network processes a batch of $M \times N$ transformed points expressed in the polygon's local frame, where $N$ represents an additional dimension, such as a receding horizon sequence along a predicted trajectory, allowing the encoder to output dual variables and distance estimates across multiple future steps. The encoder begins with a fully connected layer of size $n \times 16$, followed by layer normalization and a $\tanh$ activation. A second fully connected layer of size $16 \times 16$ with a ReLU nonlinearity is then applied. The depth of the network is constructed by repeatedly stacking these two types of layers, each maintaining 16 hidden units. The architecture concludes with a final linear layer of size $16 \times l_i$ that outputs the estimated dual variable $\hat{\vec{\mu}}_t^i$. Note that the network only predicts $\hat{\vec{\mu}}_t^i$. That is because once the dual problem converges, the corresponding dual variable $\hat{\vec{\lambda}}_t^i$ can be recovered analytically via:
\begin{equation}
\hat{\vec{\lambda}}_t^i = -\hat{\vec{\mu}}_t^i \vec{G}_i^T \vec{R}(\vec{s}_t),
\end{equation}
which follows directly from the equality constraint in the dual formulation (\ref{eq:dual_problem}). Since the encoder depends only on the polygon geometry, it only needs to be trained once for each polygon component.

For each polygon defined by $(\vec{G}_i, \vec{h}_i)$, we generate 100,000 random points within a range of $[-30,30]$ in both $x$ and $y$ directions. For each sampled point, the dual optimization problem in (\ref{eq:dual_problem_penalize}) is solved using CVXPY with the ECOS solver (\cite{domahidi2013ecos}). By slightly abusing the notation, the ECOS solver produces ground-truth optimal dual variables $\vec{\mu}_j^{i*}$, where the index $j$ denotes the $j$-th training point associated with polygon $i$. This yields a labeled dataset $\mathbb{D} = \left\{(\tilde{\vec{o}}^1,\, \vec{\mu}_1^{i*}), \dots, (\tilde{\vec{o}}^M,\, \vec{\mu}_M^{i*}) \right\}$, which is used to train the neural encoder.

The network is trained in a supervised manner to match both the dual variables and the induced signed distances. To ensure accuracy in both the vector-valued dual variables and the resulting distance expression, we use the following loss function:
\begin{equation}
\mathcal{L} =
||\hat{\vec{\mu}}_j^i - \vec{\mu}_j^{i*}||^2
+
||
{\vec{\hat{\mu}}_j^i}^T (\vec{G}_i\tilde{\vec{o}}^j - \vec{h}_i)
- {\vec{\mu}_j^{i*}}^T (\vec{G}_i\tilde{\vec{o}}^j - \vec{h}_i)
||^2
\end{equation}
where Adam optimizer is used to update the network parameters.

\section{Model Predictive Path Integral Control}
\label{sec:mppi}
Model Predictive Path Integral (MPPI) control is a sampling-based optimal control framework that solves a stochastic formulation of the finite-horizon control problem without requiring gradient information or convexity assumptions. This property makes MPPI particularly suitable for articulated agricultural vehicles, where nonlinear tractor-trailer dynamics, nonconvex obstacle geometries, and discontinuous distance-based penalties pose challenges for classical optimization-based MPC formulations.

In our formulation, the tractor–trailer dynamics is discretized as:
\begin{equation}
\vec{s}_{t+1} = \vec{s}_t + \mathcal{F}(\vec{s}_t, \vec{u}_t) \Delta t
\label{eq:mppi_dynamics}
\end{equation}
where $\Delta t$ is the sampling period and $\mathcal{F}(\cdot)$ corresponds to the continuous-time kinematic model introduced in Section~\ref{sec:kinematic_model}. To promote smoother motions and incorporate actuator limits, we augment the state with the tractor's longitudinal velocity and steering angle, and command their derivatives instead. The augmented state and control vectors become:
\begin{equation}
\vec{s}_t = [x_t, y_t, \theta_t, \phi_t, v_t, \psi_t]^T, \quad \vec{u}_t = [a_t, \zeta_t]^T
\end{equation}
where $a_t$ is longitudinal acceleration and $\zeta_t$ is the steering rate. The additional state variables evolve according to simple first-order integrators:
\begin{equation}
    v_{t+1} = v_t + a_t~\Delta t, \qquad \psi_{t+1} = \psi_t + \zeta_t~\Delta t
\end{equation}

Given a nominal control sequence $\mathbb{U}_t = \{ \vec{u}_t, \dots, \vec{u}_{t+N} \}$, where $N$ is the receding horizon length, MPPI constructs a set of perturbed control sequences by injecting Gaussian exploration noise:
\begin{equation}
\vec{u}^{(k)}_\tau = \vec{u}_\tau + \vec{w}^{(k)}_\tau, \qquad \vec{w}^{(k)}_\tau \sim \mathcal{N}(0, \vec{\Sigma}_w)
\end{equation}
for $\tau = t,\dots,t+N$. The superscript $(s)$ denotes the $k$-th sampled rollout, with $k = 1,\dots,K$ and $K$ is the number of rollouts. Each perturbed control sequence generates a simulated trajectory through tractor-trailer dynamics in (\ref{eq:mppi_dynamics}), producing $\{\vec{s}^{(k)}_{t}, \dots, \vec{s}^{(k)}_{t+N}\}$.

Each simulated trajectory is evaluated using a running cost that captures goal tracking, smooth motion, articulation feasibility, and collision avoidance. The goal tracking cost penalizes deviation from a desired goal state $\vec{s}_{goal}$:
\begin{equation}
c_{goal}(\vec{s}_{\tau}^{(k)})
= (\vec{s}_{\tau}^{(k)} - \vec{s}_{goal})^T
\mathbf{W}_g
(\vec{s}_{\tau}^{(k)} - \vec{s}_{goal}),
\end{equation}
The control cost penalizes large magnitudes of the acceleration and steering rate commands:
\begin{equation}
c_{control}(\vec{u}_{\tau}^{(k)}) = {\vec{u}_{\tau}^{(k)}}^T \vec{W}_u \vec{u}_{\tau}^{(k)}
\end{equation}
To encourage smoother motion, we penalize rapid changes in the commanded controls through:
\begin{equation}
c_{smooth}(\vec{u}_{\tau}^{(k)}, \vec{u}_{\tau-1}^{(k)}) = {(\vec{u}_{\tau}^{(k)} - \vec{u}_{\tau-1}^{(k)})}^T \vec{W}_{\Delta u} (\vec{u}_{\tau}^{(k)} - \vec{u}_{\tau-1}^{(k)})
\end{equation}

In addition, to prevent unsafe configurations, such as jacknifing on the tractor-trailer system, we include a cost term that penalizes large articulation angles:
\begin{equation}
c_{articulate}(\vec{s}_{\tau}^{(k)}) = w_{\phi}~(\phi_{\tau}^{(k)})^2
\end{equation}
A crucial component of the cost function is obstacle avoidance. At each predicted state $\vec{s}_{\tau}^{(k)}$, the minimum distance between the articulated tractor-trailer body and surrounding obstacles is computed using the geometric neural encoder introduced in Section~\ref{sec:neural_encoder}. The pretrained encoder takes LiDAR points transformed into the polygon's local frame and outputs estimated dual variables, from which accurate point-to-polygon distances can be reconstructed. Based on \eqref{eq:min_distance}, let $\
d_\tau^{(k)} = \min_{i,j} \mathcal{D}\big(\mathbb{E}^i(\vec{s}^{(k)}_\tau), \vec{o}^j_t\big)$ denote the minimum signed distance from all LiDAR points to any polygon component of the tractor-trailer geometry at time $\tau$ along the $k$-th sampled rollout. This distance is then incorporated into a smooth barrier-style potential that sharply penalizes proximity to obstacles and transitions to a linear penalty once penetration occurs:
\begin{equation}
c_{obstacle}(\vec{s}_\tau^{(k)}, d_\tau^{(k)}) =
\begin{cases}
\dfrac{w_{obs}}{d_\tau^{(k)} + \varepsilon}, & d_\tau^{(k)} > 0 \\
w_{coll}~| d_\tau^{(k)} |, & d_\tau^{(k)} \le 0
\end{cases}
\end{equation}

where $\varepsilon$ is a small positive constant. This formulation enables the controller to reason about complex articulated geometries and obstacles directly from raw sensor measurements.

Finally, to bias the final state toward the goal, we add a terminal cost:
\begin{equation}
c_{terminal}(\vec{s}_{\tau}^{(k)})
= (\vec{s}_{\tau + N}^{(k)} - \vec{s}_{goal})^T
\mathbf{W}_T
(\vec{s}_{\tau + N}^{(k)} - \vec{s}_{goal})
\end{equation}
The running cost for rollout $k$ at time $\tau$ is defined as the sum of all individual cost terms:
\begin{equation}
\begin{aligned}
\mathcal{C}\big(\vec{s}^{(k)}_\tau, \vec{u}^{(k)}_\tau, d_\tau^{(k)} \big)
=~
& c_{goal}
+ c_{control}
+ c_{smooth}
+ c_{articulate} \\
& + c_{obstacle}
+ c_{terminal}
\end{aligned}
\end{equation}
The total trajectory cost for rollout s over the prediction horizon is obtained by summing the running costs across all time steps:
\begin{equation}
J^{(k)} = \sum_{\tau=t}^{t+N} \mathcal{C}(\vec{s}^{(k)}_\tau, \vec{u}^{(k)}_\tau, d_\tau^{(k)})
\end{equation}
After that, MPPI applies an importance-sampling update inspired by path integral control theory. Each rollout $k$ is assigned a weight:
\begin{equation}
\gamma^{(k)} = \exp\!\left( -\frac{J^{(k)} - \min_j J^{(j)}}{\lambda} \right)
\end{equation}
where $\lambda > 0$ is a temperature parameter that regulates sensitivity to high-cost trajectories. The weights are normalized by:
\begin{equation}
\bar{\gamma}^{(k)} = \frac{\gamma^{(k)}}{\sum_{j=1}^{S} \gamma^{(j)}}
\end{equation}
and the nominal control sequence is updated according to a weighted average of all sampled controls:
\begin{equation}
\vec{u}_{\tau} = \sum_{s=1}^S \bar{\gamma}^{(k)} \vec{u}^{(k)}_\tau
\end{equation}
The first control input $\vec{u}_{\tau=t}$ is then applied to the tractor-trailer system, and the prediction horizon is shifted forward, yielding a receding-horizon scheme suitable for real-time use. The complete framework is summarized in Algorithm~\ref{alg:mppi}.

By combining accurate and fast neural distance estimation with sampling-based optimal control, the proposed framework generates dynamically feasible and collision-aware trajectories in cluttered and partially unknown agricultural environments. The neural encoder provides precise obstacle proximity information directly from raw LiDAR measurements, while MPPI efficiently explores perturbations that respect the nonlinear articulated dynamics and operational constraints. Together, these components form a robust and efficient navigation solution for articulated tractor-trailer system operating in complex field settings. The current framework performs goal chasing, but it can be extended to reference-path tracking by replacing $\vec{s}_{\text{goal}}$ with a time-indexed reference $\vec{s}_{\text{ref}, \tau}$ provided by a high-level planner.

\begin{algorithm}[t]
\caption{MPPI Navigation with Geometric Neural Encoder}
\label{alg:mppi}
\begin{spacing}{1.0}
\begin{algorithmic}[1]
\State Initialize state $\vec{s}_0$, control sequence $\mathbb{U}_0=\{\vec{u}_0,\dots,\vec{u}_N\}$, set $t\gets 0$
\While{task not finished}
    \State Acquire LiDAR scan and form obstacle set $\mathbb{O}_t$
    \For{$k=1,\dots,K$}
        \State $\vec{s}^{(k)}_t \gets \vec{s}_t$
        \For{$\tau=t,\dots,t+N$}
            \State Sample $\vec{u}^{(k)}_\tau = \vec{u}_\tau + \vec{w}^{(k)}_\tau$, $\vec{w}^{(k)}_\tau\!\sim\!\mathcal{N}(0,\vec{\Sigma}_w)$
            \State Propagate: $\vec{s}^{(k)}_{\tau+1} = \vec{s}^{(k)}_\tau + \mathcal{F}(\vec{s}^{(k)}_\tau,\vec{u}^{(k)}_\tau)\Delta t$
            \State Compute distance $d_\tau^{(k)}$ via neural encoders
            \State Evaluate running cost $\mathcal{C}(\vec{s}^{(k)}_\tau,\vec{u}^{(k)}_\tau, d_\tau^{(k)})$
        \EndFor
        \State Compute total cost $J^{(k)}$
    \EndFor
    \State Compute normalized weights $\bar{\gamma}^{(k)}$ from $J^{(k)}$
    \For{$\tau=t,\dots,t+N$}
        \State Update control: $\vec{u}_\tau = \sum_s \bar{\gamma}^{(k)}\,\vec{u}^{(k)}_\tau$
    \EndFor
    \State Apply $\vec{u}_t$, observe $\vec{s}_{t+1}$, shift horizon, set $t\gets t+1$
\EndWhile
\end{algorithmic}
\end{spacing}
\end{algorithm}

\section{Simulation Results}
\label{sec:results}
We evaluate the performance of the proposed navigation framework in a simulated environment containing dense obstacles and narrow free-space passages. The articulated tractor–trailer model used in simulation consists of three convex polygons: one for the tractor body and two for the trailer, as shown in Fig.~\ref{fig:tractor_trailer_diagram}a. The tractor measures $3.35\,\mathrm{m} \times 1.48\,\mathrm{m}$, while the trailer includes a rectangular body of $1.2\,\mathrm{m} \times 3.6\,\mathrm{m}$ and a triangular connector part. The tractor wheelbase is $L_0 = 1.9\,\mathrm{m}$, the hitch offset is $L_h = 0.5\,\mathrm{m}$, and the trailer axle is located $L_1 = 1.5\,\mathrm{m}$ behind the hitch point.

A geometric neural encoder is trained for each convex polygon. Training all three encoders requires approximately 2.2 hours on a machine equipped with an Intel i9-13900KF CPU and an NVIDIA RTX 4090 GPU. Each encoder is trained for 5,000 epochs using 80/20 train-test split of the dataset. After training, the mean squared error (MSE) between predicted and ground-truth dual variables and point-to-polygon distances on the test set drops below $1 \times 10^{-5}$, indicating strong convergence. Once trained, the encoders are integrated into the MPPI controller and used online without further optimization.

A simulated 2D LiDAR sensor provides point clouds at every time step, which serve as input to the neural encoders. The robot’s state is assumed to be known; on real platforms, this could be provided by a localization module running a Simultaneous Localization and Mapping (SLAM) algorithm. The MPPI controller is configured with a prediction horizon of 50 steps, a time step of $\Delta t = 0.1\,\mathrm{s}$, 1000 sampled trajectories per update, Gaussian exploration noise $\Sigma_w = \mathrm{diag}(2, 2)$, and a temperature parameter $\lambda = 1$. The running cost uses the following weights: $W_g=\mathrm{diag}(1, 1, 0.5, 0.5, 0, 0)$ for goal tracking, $W_u = \mathrm{diag}(0.1,0.1)$ for control effort, $W_{\Delta u} = \mathrm{diag}(0.1,0.1)$ for smoothness, $w_\phi = 1$ for articulation penalty, and $w_{obs} = 5$, $w_{coll} = 50$ for collision. A terminal cost with weight $W_T = 10~W_g$ is applied to bias the final state toward the goal. These parameters are tuned empirically. The controller is implemented in Python with GPU acceleration.

\begin{figure}[t!]
\centering
\includegraphics[width=0.65\linewidth]{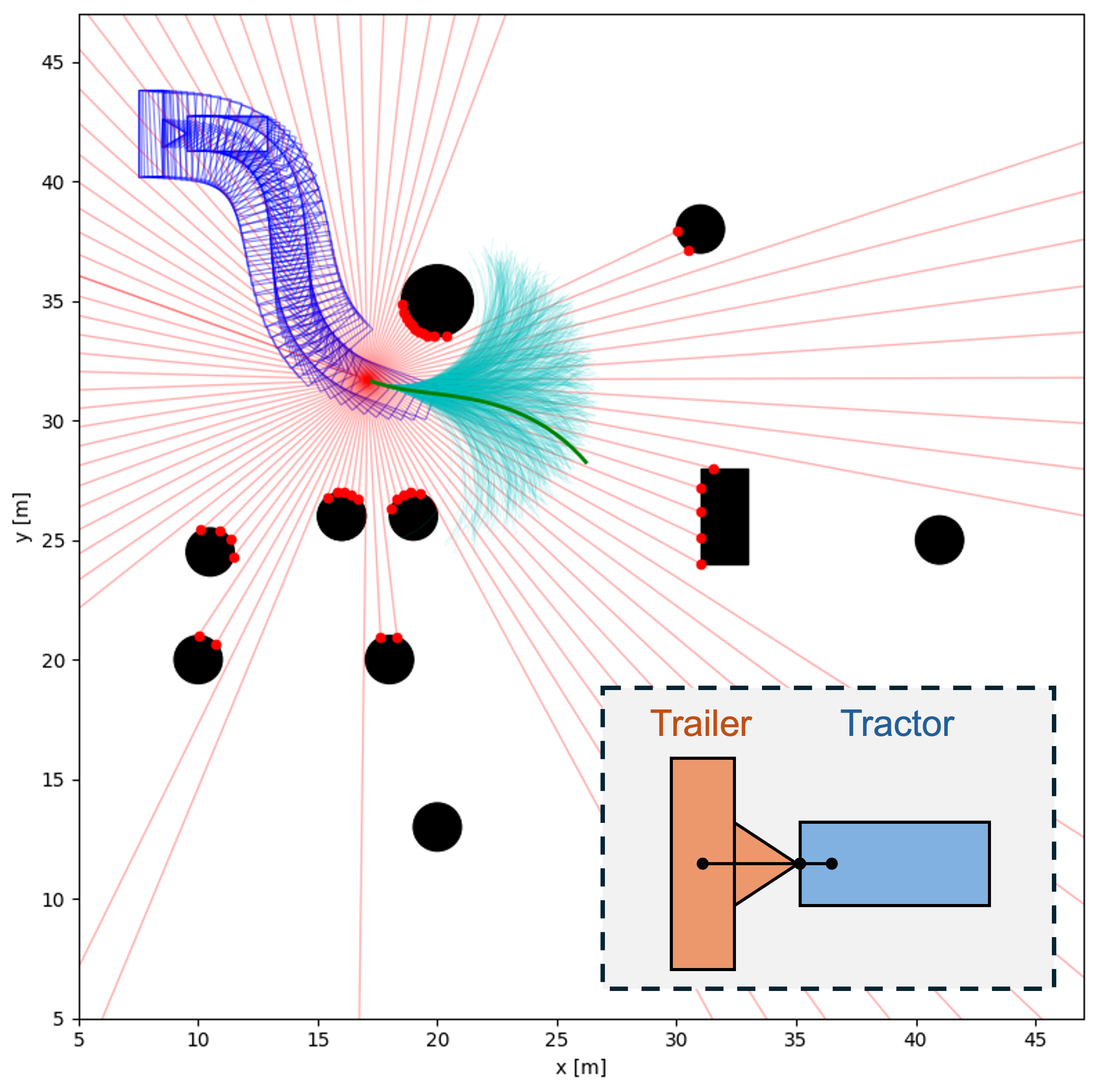}
\includegraphics[width=0.65\linewidth]{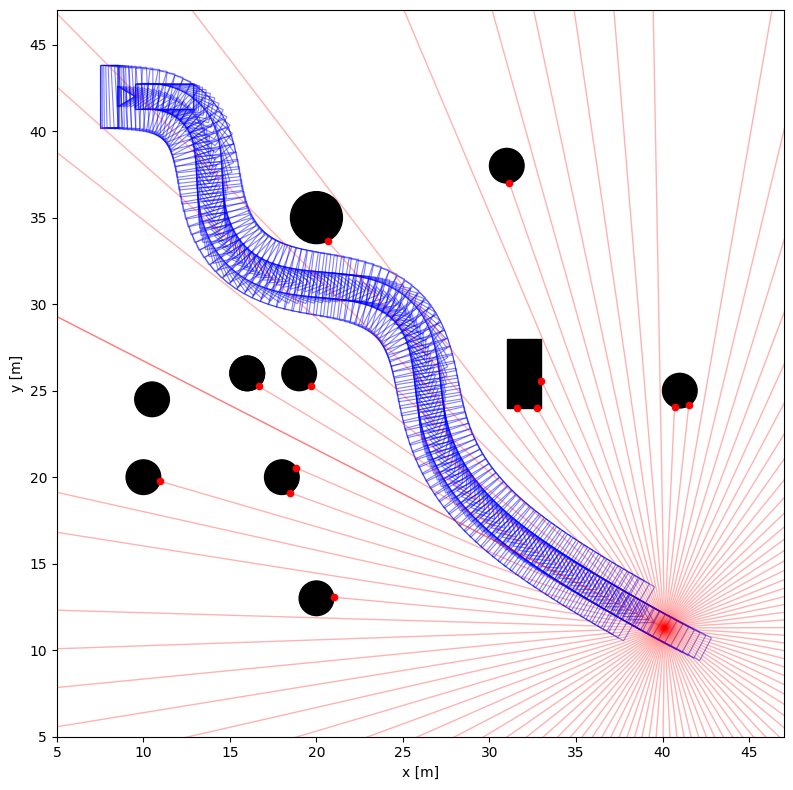}
\caption{MPPI-based navigation of a tractor–trailer: sampled rollouts and optimal trajectory (top), final collision-free arrival at the goal (bottom).}
\label{fig:navigation}
\end{figure}

Figure~\ref{fig:navigation} illustrates the navigation performance of the proposed system. In Fig.~\ref{fig:navigation}a, the tractor–trailer starts from the upper-left corner at $\vec{s}_{start} = [10, 42, 0, 0, 0, 0]^T$ and needs to reach a goal in the lower-right region at $\vec{s}_{goal} = [40, 12, -0.5, 0, 0, 0]^T$ while avoiding a dense field of obstacles. At each control step, the MPPI controller samples candidate trajectories and selects the optimal one based on the neural distance-informed cost described in Algorithm \ref{alg:mppi}. Because all sampled rollouts are generated using the tractor–trailer kinematic model, dynamic feasibility is naturally enforced throughout the process. The rendered trajectory history shows that the tractor–trailer maintains safe clearance throughout the maneuver without contacting any obstacles. The configuration in Fig.\ref{fig:navigation}b confirms the successful arrival at the goal. The controller operates at approximately 30 Hz in simulation, demonstrating real-time feasibility. These results demonstrate the effectiveness of the proposed neural distance–guided MPPI control framework for safe, articulated tractor–trailer navigation in cluttered environments.

\section{Conclusion and Future Work}
\label{sec:conclusion}
This work presented a neural distance–guided Model Predictive Path Integral control framework for real-time navigation of articulated tractor–trailer systems in cluttered environments. By combining a duality-inspired geometric neural encoder with a sampling-based stochastic control method, the proposed pipeline achieves accurate distance estimation and safe, dynamically feasible navigation for tractor-trailer systems without requiring prior maps or geometric simplifications.

While effective, the current framework requires training a separate neural encoder for each polygon geometry. A promising direction for future work is the development of a universal or geometry-conditioned encoder that can generalize across multiple tractor–trailer configurations without retraining. Since MPPI functions as a local planner, it may become trapped in local minima in highly constrained environments, and integrating our approach with a global planner would provide high-level guidance and improve robustness. Real-world deployment is another important next step, which will require filtering ground returns and overhanging obstacles by selecting 3D LiDAR points within a height band of interest and projecting them onto the 2D plane, together with outlier rejection for sensor noise. The neural encoder processes points in parallel on the GPU and scales favorably with point count, and voxel downsampling can further reduce input size for embedded platforms. Additional future work includes expanded studies across diverse field layouts, more extensive testing in reverse-motion scenarios, evaluation under dynamic and moving obstacles, and quantitative comparisons with state-of-the-art planners.

% \begin{ack}
% % Place acknowledgments here.
% \end{ack}

% \section*{DECLARATION OF GENERATIVE AI AND AI-ASSISTED TECHNOLOGIES IN THE WRITING PROCESS}
% During the preparation of this work the author(s) used [NAME TOOL / SERVICE] in order to [REASON]. After using this tool/service, the author(s) reviewed and edited the content as needed and take(s) full responsibility for the content of the publication.

\bibliography{ifacconf}             % bib file to produce the bibliography
                                                     % with bibtex (preferred)

% \appendix
% \section{A summary of Latin grammar}    % Each appendix must have a short title.
% \section{Some Latin vocabulary}              % Sections and subsections are supported  
                                                                         % in the appendices.
\end{document}